\documentclass[11pt]{article}
\usepackage[margin=1in]{geometry}
\usepackage{amsmath,amssymb,amsthm,bm,graphicx,booktabs,microtype}
\usepackage[numbers,sort&compress]{natbib}
\usepackage{hyperref}
\hypersetup{colorlinks=true,linkcolor=blue,citecolor=blue,urlcolor=blue}
\newtheorem{proposition}{Proposition}

\title{PEEL: Physics-Enabled Evidential Learning for Identifiable Uncertainty in CT Imaging}
\author{Ge Wang\\Biomedical Imaging Center, Rensselaer Polytechnic Institute\\Troy, New York, USA}
\date{August 27, 2026}

\begin{document}
\maketitle

\begin{abstract}
Normal--inverse-gamma (NIG) regression is not uniquely identifiable from its marginal Student-$t$ likelihood: the likelihood determines three combinations of four NIG parameters and is constant along a one-dimensional fiber. We identify that fiber using independent physical measurement. As an initial embodiment, a reconstruction network receives one noisy filtered-backprojection (FBP) image and is first trained only by Student-$t$ negative log-likelihood to estimate the three identifiable coordinates $(\gamma,\alpha,c)$. The network is then frozen; repeated physical-noise realizations propagated through its reconstruction output form a Monte Carlo (MC) teacher label for output-domain aleatoric variance. An aleatoric head attached to frozen features learns this label, after which $(\beta,\nu)$ are recovered algebraically. On 30 held-out simulated objects at five photon levels, one-image predictions achieved pooled Spearman correlations of 0.832--0.951 against independent 400-repeat references, median within-image correlations were 0.834--0.947, and 98.81--99.55\% of evaluated pixels satisfied the algebraic admissibility condition. The method needs no KL term, reference prior, evidence regularizer, or cross-loss weight.
\end{abstract}

\noindent\textbf{Keywords:}
Physics-enabled evidential learning (PEEL); uncertainty quantification;
normal-inverse-gamma regression; identifiability; aleatoric uncertainty;
computed tomography (CT); Monte Carlo simulation.

\section{Introduction}
Normal–inverse-gamma (NIG) regression ~\cite{amini2020deep} provides a hierarchical probabilistic model for predictive uncertainty. The scientific question is how to resolve NIG non-identifiability using new information rather than a selected regularizer or reference prior. The proposed answer is
\begin{equation}
\boxed{
\begin{gathered}
\text{NLL learns }(\gamma,\alpha,c)\text{ and reconstruction features},\\
\text{MC supervises an aleatoric head},\qquad
\text{algebra recovers }(\beta,\nu).
\end{gathered}}
\label{eq:statement}
\end{equation}
The first and second lines are optimized sequentially, never as a weighted sum. The frozen features from the first line are anatomical and noise-sensitive information already learned by the reconstruction network and allow the aleatoric head to learn effectively, without allowing the aleatoric loss to alter the predictive model.

Deep evidential regression commonly supplements the marginal likelihood with evidence regularization \citep{amini2020deep,meinert2023unreasonable}, while prior-network approaches specify a reference distribution \citep{malinin2018prior}. Our earlier ELVAE formulation placed an input-dependent NIG hierarchy at a variational latent bottleneck and explicitly distinguished the hierarchy-level uncertainty decomposition from the marginalized Student-$t$ law \citep{wang2026elvae}. Such regularized hierarchical models may select a point on an unidentified fiber, but the regularizer does not constitute an additional physical observation. The present MC teacher is intended to supply that missing observation directly in CT reconstruction.

The remainder of this paper is organized as follows. Section 2 presents the PEEL framework, beginning with the NIG parameterization and its non-identifiability, followed by the CT simulation model, Student-$t$ NLL training, Monte Carlo supervision of output-domain aleatoric variance, and algebraic recovery of the full NIG parameters. Section 3 describes the experimental design and evaluates single-image aleatoric-variance prediction against independent 400-repeat Monte Carlo references across five photon levels. Section 4 summarizes the main findings, limitations, and implications of physics-enabled identification for evidential learning.

\section{Methodology}
\begin{figure}[ht]
\centering
\includegraphics[width=\linewidth]{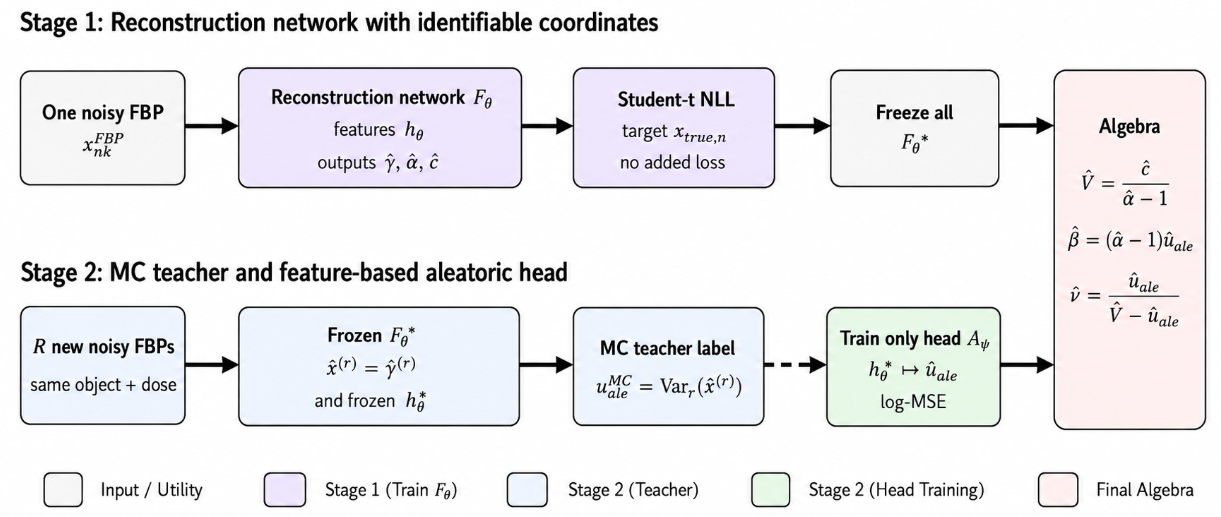}
\caption{PEEL workflow with a sequential teacher--student design. Stage~1 trains $F_\theta$ from noisy FBP using only Student-$t$ NLL. Repeated physical-noise inputs propagated through frozen $F_{\theta^\star}$ generate $u_{\mathrm{ale}}^{\mathrm{MC}}$. Stage~2 attaches $A_\psi$ to frozen features $h_{\theta^\star}$ and updates only $\psi$.}
\label{fig:design}
\end{figure}

Figure~\ref{fig:design} illustrates the overall workflow of the proposed Physics-Enabled Evidential Learning (PEEL) framework. The method separates likelihood-based estimation from physics-based uncertainty identification by first training and freezing the reconstruction network, and then learning output-domain aleatoric variance from repeated physical-noise realizations. The resulting likelihood parameters and independently estimated aleatoric variance are finally combined through closed-form algebra to recover the full NIG parameterization.

\subsection{NIG parameterization and identifiability}
At pixel $i$, the NIG hierarchy is
\begin{align}
\sigma_i^2 &\sim \mathrm{InvGamma}(\alpha_i,\beta_i),\\
\mu_i\mid\sigma_i^2 &\sim \mathcal N\!\left(\gamma_i,\frac{\sigma_i^2}{\nu_i}\right),\\
z_i\mid\mu_i,\sigma_i^2 &\sim \mathcal N(\mu_i,\sigma_i^2),
\end{align}
with $\alpha_i>1$, $\beta_i>0$, and $\nu_i>0$. Its marginal law is
\begin{equation}
z_i\sim t_{2\alpha_i}\!\left(\gamma_i,\frac{c_i}{\alpha_i}\right),
\qquad
c_i=\beta_i\left(1+\frac1{\nu_i}\right).
\label{eq:marginal}
\end{equation}
The marginal predictive variance is
\begin{equation}
V_{\mathrm{pred},i}=\frac{c_i}{\alpha_i-1}.
\label{eq:vpred}
\end{equation}
The NLL identifies $(\gamma_i,\alpha_i,c_i)$ but not $(\beta_i,\nu_i)$ separately, because every
\begin{equation}
\beta_i(\nu_i)=\frac{c_i\nu_i}{1+\nu_i},\qquad \nu_i>0,
\label{eq:fiber}
\end{equation}
gives the same marginal Student-$t$ distribution.

The NIG decomposition is
\begin{equation}
u_{\mathrm{ale},i}=\frac{\beta_i}{\alpha_i-1},
\qquad
u_{\mathrm{epi},i}=\frac{\beta_i}{\nu_i(\alpha_i-1)},
\qquad
V_{\mathrm{pred},i}=u_{\mathrm{ale},i}+u_{\mathrm{epi},i}.
\label{eq:decomp}
\end{equation}

\begin{proposition}
If $(\gamma_i,\alpha_i,c_i)$ and an independent $u_{\mathrm{ale},i}$ are known, with
\begin{equation}
0<u_{\mathrm{ale},i}<\frac{c_i}{\alpha_i-1},
\end{equation}
then the unique NIG representative is
\begin{equation}
\boxed{\beta_i=(\alpha_i-1)u_{\mathrm{ale},i}},
\qquad
\boxed{\nu_i=\frac{u_{\mathrm{ale},i}}
{c_i/(\alpha_i-1)-u_{\mathrm{ale},i}}}.
\label{eq:recover}
\end{equation}
\end{proposition}
No KL term or reference prior appears in Eq.~\eqref{eq:recover}.

\subsection{CT data generation}
The initial proof-of-concept uses:
\begin{center}
\begin{tabular}{ll}
\toprule
Quantity & Fixed value\\
\midrule
Image size & $32\times32$ pixels\\
Parallel-beam views & 36 over $[0,180^\circ)$\\
Objects (training, validation, and testing) & $N_{\mathrm{tr}}=100$, $N_{\mathrm{va}}=20$, $N_{\mathrm{te}}=30$\\
Dose levels & $L=5$\\
NLL realizations per training object-dose pair & $K_{\mathrm{NLL}}=4$\\
Aleatoric-input realizations per training pair & $K_{\mathrm{ale}}=4$\\
Teacher repeats & $R_{\mathrm{teach}}=32$\\
Reference repeats & $R_{\mathrm{ref}}=400$\\
Electronic-noise SD & $\sigma_e=60$ counts\\
\bottomrule
\end{tabular}
\end{center}

Indices are:
\begin{equation}
n=1,\ldots,N \quad\text{(object)},\qquad
\ell=1,\ldots,L \quad\text{(dose)},
\end{equation}
\begin{equation}
k=1,\ldots,K \quad\text{(single-input realization)},\qquad
r=1,\ldots,R \quad\text{(MC repeat)},
\end{equation}
and $i\in\Omega$, where $|\Omega|=32^2=1024$. Objects, not noise realizations, are separated across training, validation, and test splits.

Each $32\times32$ object contains a random body ellipse with attenuation in $[0.017,0.022]$, four to seven internal ellipses with signed contrast in $[-0.005,0.006]$, and one to three small high-contrast lesions with added attenuation in $[0.006,0.011]$. Values are clipped to $[0,0.035]$. Independent random seeds generate the 100/20/30 training/validation/testing objects.

For an object $n$, let its ground-truth image be $x_n$ and
\begin{equation}
p_{nj}=(Ax_n)_j
\label{eq:projection}
\end{equation}
be its noiseless line integral at detector-view bin $j$. Every object is simulated at the five fixed photon levels
\begin{equation}
I_{0,\ell}\in
\{1.50,\;2.25,\;3.50,\;5.25,\;8.00\}\times10^4.
\label{eq:doses}
\end{equation}
For realization $q$ at object $n$ and dose $\ell$,
\begin{align}
N_{n\ell qj}&\sim\mathrm{Poisson}
\left(I_{0,\ell}e^{-p_{nj}}\right),\\
e_{n\ell qj}&\sim\mathcal N(0,\sigma_e^2),\\
\widetilde N_{n\ell qj}&=\max(N_{n\ell qj}+e_{n\ell qj},0.5),\\
y_{n\ell qj}&=-\log\left(\frac{\widetilde N_{n\ell qj}}{I_{0,\ell}}\right),\\
x^{\mathrm{FBP}}_{n\ell q}&=\operatorname{FBP}(y_{n\ell q}).
\label{eq:fbp}
\end{align}
The 0.5-photon floor prevents non-positive noisy counts before the logarithmic transform while introducing only a minimal half-count correction in the extreme low-count regime.
All Poisson and Gaussian draws are independent across $(n,\ell,q,j)$. The network input is only $x^{\mathrm{FBP}}$; $I_0$, the sinogram, and the clean image are not input channels. Each training object appears at every dose, and object-dose pairs are sampled uniformly.

\subsection{Stage 1: reconstruction network under NLL}
For NLL realization $k$,
\begin{equation}
(h_{n\ell k},\widehat\gamma_{n\ell k},
\widehat\alpha_{n\ell k},\widehat c_{n\ell k})
=F_\theta(x^{\mathrm{FBP}}_{n\ell k}),
\label{eq:gnet}
\end{equation}
where $h_{n\ell k}$ is the shared feature tensor and
$(\widehat\gamma,\widehat\alpha,\widehat c)$ are three image-valued outputs. At pixel $i$, positivity is enforced by
\begin{align}
\widehat\alpha_{n\ell ki}
&=1+\operatorname{softplus}\!\left((W_\alpha\ast h_{n\ell k})_i\right)+\epsilon_\alpha,
&\widehat c_{n\ell ki}
&=s^{-2}\operatorname{softplus}\!\left((W_c\ast h_{n\ell k})_i\right)+\epsilon_0,\\
s&=50,
&\epsilon_\alpha&=10^{-4},\qquad\epsilon_0=10^{-8},
\end{align}
where $W_\alpha$ and $W_c$ denote the corresponding $1\times1$ convolutional heads. The fixed scale
$s$ is used to keep $c$, which has squared-intensity units, in a well-conditioned range during optimization; accordingly, the factor $s^{-2}$ rescales the positive network output to the appropriate scale.
The clean target for every noise realization of object $n$ follows the pixelwise model:
\begin{equation}
x_{ni}\mid x^{\mathrm{FBP}}_{n\ell k}
\sim
t_{2\widehat\alpha_{n\ell ki}}
\left(\widehat\gamma_{n\ell ki},
\frac{\widehat c_{n\ell ki}}{\widehat\alpha_{n\ell ki}}\right).
\end{equation}
Define $e_{n\ell ki}=x_{ni}-\widehat\gamma_{n\ell ki}$. The exact pixelwise NLL is
\begin{align}
\ell^{\mathrm{NLL}}_{n\ell ki}
={}&\log\Gamma(\widehat\alpha_{n\ell ki})
-\log\Gamma\left(\widehat\alpha_{n\ell ki}+\tfrac12\right)
+\tfrac12\log(2\pi\widehat c_{n\ell ki})\nonumber\\
&+\left(\widehat\alpha_{n\ell ki}+\tfrac12\right)
\log\left(1+\frac{e_{n\ell ki}^2}{2\widehat c_{n\ell ki}}\right).
\label{eq:nllpixel}
\end{align}
The complete loss is
\begin{equation}
\boxed{
\mathcal J_{\mathrm{NLL}}(\theta)
=\frac{1}{N_{\mathrm{tr}}LK_{\mathrm{NLL}}|\Omega|}
\sum_{n=1}^{N_{\mathrm{tr}}}
\sum_{\ell=1}^{L}
\sum_{k=1}^{K_{\mathrm{NLL}}}
\sum_{i\in\Omega}
\ell^{\mathrm{NLL}}_{n\ell ki}.}
\label{eq:nllfull}
\end{equation}
The network parameters are optimized for
\begin{equation}
\theta^\star=\arg\min_\theta\mathcal J_{\mathrm{NLL}}(\theta).
\end{equation}
Then, all the parameters of $F_{\theta^\star}$ are frozen. There is no additional $\gamma$ loss, KL term, evidence penalty, or aleatoric loss in Stage~1.

For a single input, Stage~1 gives
\begin{equation}
\widehat V_{\mathrm{pred},n\ell ki}
=\frac{\widehat c_{n\ell ki}}{\widehat\alpha_{n\ell ki}-1}.
\label{eq:vhat}
\end{equation}

For the proof-of-concept, $F_\theta$ scales its FBP input by $s=50$, then applies a $3\times3$ convolution from one input channel to 32 feature channels followed by four residual blocks. Each block contains two $3\times3$ convolutions with 32 channels and a ReLU between them. The final tensor is $h_\theta$. Three separate $1\times1$ convolutions produce a reconstruction $\hat \gamma$, $\hat \alpha$, and $\hat c$:
\begin{equation}
\widehat\gamma=x^{\mathrm{FBP}}+s^{-1}W_\gamma*h_\theta,
\quad
\widehat\alpha=1+\operatorname{softplus}(W_\alpha*h_\theta)+\epsilon_\alpha,
\quad
\widehat c=s^{-2}\operatorname{softplus}(W_c*h_\theta)+\epsilon_0.
\label{eq:architecture}
\end{equation}
The aleatoric head $A_\psi$ contains one $3\times3$ convolution with 32 channels and ReLU, followed by a $1\times1$ convolution and $s^{-2}\operatorname{softplus}(\cdot)+\epsilon_0$. It receives $h_{\theta^\star}$, not an independently reconstructed image. The scale $s$ changes numerical units inside the network but not the physical output units. These choices are fixed for reproducibility.

\subsection{Stage 2: Monte Carlo teacher and aleatoric head}
For each object-dose pair $(n,\ell)$, generate a teacher pool using noise seeds disjoint from Stage~1. Pass every repeated FBP through frozen $F_{\theta^\star}$ and retain only its reconstruction output:
\begin{equation}
\widehat x^{(r)}_{n\ell i}
=\widehat\gamma_{\theta^\star}
\left(x^{\mathrm{FBP},(r)}_{n\ell}\right)_i,
\qquad r=1,\ldots,R_{\mathrm{teach}}.
\label{eq:mcout}
\end{equation}
Let us define
\begin{equation}
\overline x_{n\ell i}
=\frac1{R_{\mathrm{teach}}}
\sum_{r=1}^{R_{\mathrm{teach}}}\widehat x^{(r)}_{n\ell i}
\end{equation}
and compute the output-domain label as follows:
\begin{equation}
\boxed{
u^{\mathrm{MC}}_{\mathrm{ale},n\ell i}
=\frac1{R_{\mathrm{teach}}-1}
\sum_{r=1}^{R_{\mathrm{teach}}}
\left(\widehat x^{(r)}_{n\ell i}-\overline x_{n\ell i}\right)^2.}
\label{eq:ualelabel}
\end{equation}

Generate $K_{\mathrm{ale}}$ additional FBP realizations, disjoint from the NLL and teacher pools. For each realization, compute the frozen feature tensor
\begin{equation}
h^\star_{n\ell k}=h_{\theta^\star}(x^{\mathrm{FBP}}_{n\ell k}).
\end{equation}
The trainable aleatoric head is
\begin{equation}
\widehat u_{\mathrm{ale},n\ell ki}
=A_\psi(h^\star_{n\ell k})_i>0.
\end{equation}
Using a fixed numerical floor $\epsilon_0=10^{-8}$, its only loss is
\begin{equation}
\boxed{
\mathcal J_{\mathrm{ale}}(\psi)
=\frac{1}{N_{\mathrm{tr}}LK_{\mathrm{ale}}|\Omega|}
\sum_{n=1}^{N_{\mathrm{tr}}}
\sum_{\ell=1}^{L}
\sum_{k=1}^{K_{\mathrm{ale}}}
\sum_{i\in\Omega}
\left[
\log(\widehat u_{\mathrm{ale},n\ell ki}+\epsilon_0)
-\log(u^{\mathrm{MC}}_{\mathrm{ale},n\ell i}+\epsilon_0)
\right]^2.}
\label{eq:aleloss}
\end{equation}
Naturally, this aleatoric head can be optimized for
\begin{equation}
\psi^\star=\arg\min_\psi\mathcal J_{\mathrm{ale}}(\psi).
\end{equation}
Stages 1 and 2 perform optimizations with separate losses. During Stage~2, $\theta^\star$ is fixed and gradients update only $\psi$. Since the losses are never added, there is no cross-loss coefficient. 
Note that the``teacher'' denotes the Monte Carlo target in Eq.~\eqref{eq:ualelabel}, which is not a second trainable teacher network.

Both stages use AdamW with batch size 32, learning rate $10^{-3}$, weight decay $10^{-5}$, and gradient-norm clipping at 5. Training runs for at most 80 epochs; the checkpoint with the lowest validation loss is retained, with early stopping after 12 unimproved epochs once at least 25 epochs have elapsed. The reported proof-of-concept uses the fixed seed 20260823. Stage~1 uses $K_{\mathrm{NLL}}=4$ inputs per training object-dose pair, and Stage~2 uses $K_{\mathrm{ale}}=4$ different inputs plus the disjoint $R_{\mathrm{teach}}=32$ teacher pool.

\subsection{Stage 3: algebraic recovery}
For one test FBP input $x^{\mathrm{FBP}}$, compute
\begin{equation}
(h^\star,\widehat\gamma,\widehat\alpha,\widehat c)
=F_{\theta^\star}(x^{\mathrm{FBP}}),
\qquad
\widehat u_{\mathrm{ale}}=A_{\psi^\star}(h^\star),
\end{equation}
and
\begin{equation}
\widehat V_{\mathrm{pred}}=\frac{\widehat c}{\widehat\alpha-1}.
\end{equation}
For pixels satisfying
\begin{equation}
0<\widehat u_{\mathrm{ale}}<\widehat V_{\mathrm{pred}},
\label{eq:admissible}
\end{equation}
recover
\begin{equation}
\boxed{\widehat\beta=(\widehat\alpha-1)\widehat u_{\mathrm{ale}}},
\qquad
\boxed{\widehat\nu=\frac{\widehat u_{\mathrm{ale}}}
{\widehat V_{\mathrm{pred}}-\widehat u_{\mathrm{ale}}}}.
\label{eq:plugin}
\end{equation}
Pixels violating Eq.~\eqref{eq:admissible} are reported as failures.

\begin{figure}[ht]
\centering
\includegraphics[width=\linewidth]{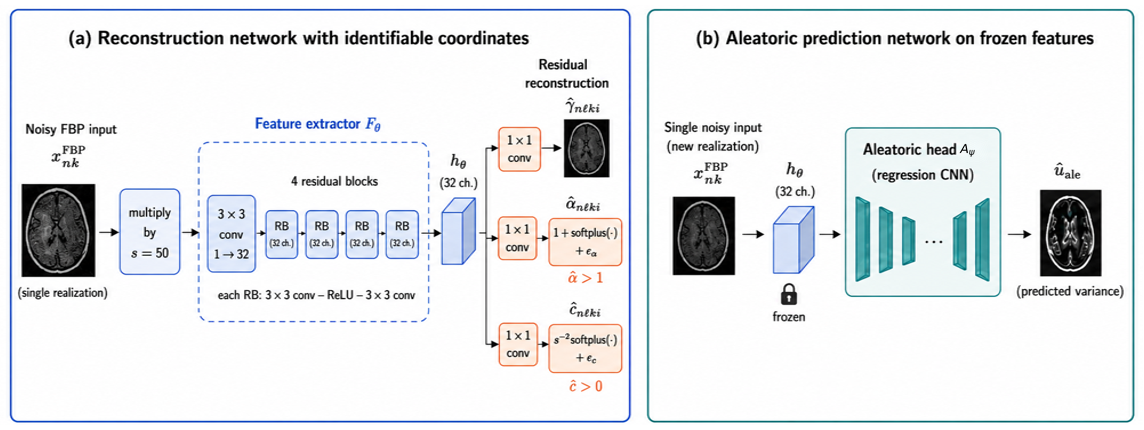}
\caption{Two-network architectures for PEEL. (a) In Stage~1, the reconstruction network $F_\theta$ maps a single noisy FBP realization to shared features $h_\theta$ and the identifiable Student-$t$ coordinates $(\hat{\gamma},\hat{\alpha},\hat{c})$. (b) In Stage~2, with $F_\theta$ frozen, the aleatoric head $A_\psi$ uses $h_\theta$ from a single noisy realization to predict the output-domain aleatoric variance $\hat{u}_{\mathrm{ale}}$, supervised by the Monte Carlo teacher $u_{\mathrm{ale}}^{\mathrm{MC}}$ constructed from repeated physical-noise realizations. The decoupled design avoids cross-loss coupling between reconstruction and aleatoric regression.}
\label{Network}
\end{figure}

Figure~\ref{Network} summarizes the two-network architectures, highlighting the separation between Student-$t$ distribution-based reconstruction in Stage~1 and measurement-anchored aleatoric prediction in Stage~2.

\section{Experimental Design and Results}
\subsection{Overall design}
The primary question here is whether $A_{\psi^\star}$ can infer the output-domain aleatoric variance from one noisy FBP for a previously unseen object drawn from the same distribution as the training objects. No out-of-distribution claim is made.

For every held-out test object $n=1,\ldots,N_{\mathrm{te}}$ and dose $\ell=1,\ldots,L$, generate one evaluation input $x^{\mathrm{FBP},(0)}_{n\ell}$ and an independent reference pool $\{x^{\mathrm{FBP},(r)}_{n\ell}\}_{r=1}^{R_{\mathrm{ref}}}$, where $R_{\mathrm{ref}}=400$. Neither the evaluation input nor any reference realization is used in training process. From the single input,
\begin{equation}
(h^{(0)}_{n\ell},\widehat\gamma^{(0)}_{n\ell},
\widehat\alpha^{(0)}_{n\ell},\widehat c^{(0)}_{n\ell})
=F_{\theta^\star}(x^{\mathrm{FBP},(0)}_{n\ell}),
\qquad
\widehat u^{(0)}_{\mathrm{ale},n\ell}
=A_{\psi^\star}(h^{(0)}_{n\ell}).
\label{eq:singleprediction}
\end{equation}
The independent MC reference is obtained by propagating all $R_{\mathrm{ref}}$ realizations through the frozen reconstruction output:
\begin{align}
\widehat x^{(r)}_{n\ell i}
&=\widehat\gamma_{\theta^\star}
\left(x^{\mathrm{FBP},(r)}_{n\ell}\right)_i,\\
\overline x^{\mathrm{ref}}_{n\ell i}
&=\frac1{R_{\mathrm{ref}}}
\sum_{r=1}^{R_{\mathrm{ref}}}\widehat x^{(r)}_{n\ell i},\\
u^{\mathrm{ref}}_{\mathrm{ale},n\ell i}
&=\frac1{R_{\mathrm{ref}}-1}
\sum_{r=1}^{R_{\mathrm{ref}}}
\left(\widehat x^{(r)}_{n\ell i}
-\overline x^{\mathrm{ref}}_{n\ell i}\right)^2.
\label{eq:reference}
\end{align}
Thus, Eq.~\eqref{eq:singleprediction} uses one image, whereas Eq.~\eqref{eq:reference} is used only to evaluate that prediction.

For object $n$, let $\Omega_b^{(n)}$ be the known nonzero support of $x_n$, morphologically eroded by three pixels. All pixelwise metrics use $i\in\Omega_b^{(n)}$. With
\begin{equation}
M_\ell=\sum_{n=1}^{N_{\mathrm{te}}}|\Omega_b^{(n)}|,
\end{equation}
the dose-specific log-RMSE is
\begin{equation}
\operatorname{logRMSE}_\ell=
\left\{
\frac1{M_\ell}
\sum_{n=1}^{N_{\mathrm{te}}}
\sum_{i\in\Omega_b^{(n)}}
\left[
\log(\widehat u^{(0)}_{\mathrm{ale},n\ell i}+\epsilon_0)
-\log(u^{\mathrm{ref}}_{\mathrm{ale},n\ell i}+\epsilon_0)
\right]^2
\right\}^{1/2}.
\label{eq:logrmse}
\end{equation}
For each dose, we also report (i) pooled Spearman correlation over all test-mask pixels; (ii) the median of within-image Spearman correlations; and (iii) case-level Spearman correlation between the spatial means
\begin{equation}
\overline{\widehat u}^{(0)}_{n\ell}
=\frac1{|\Omega_b^{(n)}|}\sum_{i\in\Omega_b^{(n)}}
\widehat u^{(0)}_{\mathrm{ale},n\ell i},
\qquad
\overline u^{\mathrm{ref}}_{n\ell}
=\frac1{|\Omega_b^{(n)}|}\sum_{i\in\Omega_b^{(n)}}
u^{\mathrm{ref}}_{\mathrm{ale},n\ell i}.
\label{eq:casemeans}
\end{equation}

Because algebraic NIG recovery additionally requires Eq.~\eqref{eq:admissible}, we report the dose-specific admissible fraction
\begin{equation}
q_\ell=
\frac1{M_\ell}
\sum_{n=1}^{N_{\mathrm{te}}}
\sum_{i\in\Omega_b^{(n)}}
\mathbf 1\!\left[
0<\widehat u^{(0)}_{\mathrm{ale},n\ell i}
<\frac{\widehat c^{(0)}_{n\ell i}}
{\widehat\alpha^{(0)}_{n\ell i}-1}
\right].
\label{eq:admissiblefraction}
\end{equation}
\medskip
No separate ground truth is assigned to $u_{\mathrm{epi}}$, $\nu$, or $\beta$. Their values are algebraic consequences of the learned likelihood coordinates and the independently supervised $u_{\mathrm{ale}}$.

\subsection{Representative results}
Table~\ref{tab:results} reports the complete predeclared experiment. The aleatoric head recovered spatial ordering consistently at all five doses: pooled $\rho$ increased from 0.832 at $I_0=1.5\times10^4$ to 0.951 at $8.0\times10^4$, while median within-image $\rho$ increased from 0.834 to 0.947. Case-mean correlations were between 0.819 and 0.886. Log-RMSE was lowest at the middle dose, indicating that spatial ordering was more stable than absolute log-scale calibration at the dose extremes. The admissible fraction exceeded 98.8\% at every dose.

\begin{table}[t]
\centering
\caption{One-image $u_{\mathrm{ale}}$ prediction against independent $R_{\mathrm{ref}}=400$ MC references on 30 held-out objects. All correlations are Spearman coefficients.}
\label{tab:results}
\small
\begin{tabular}{rrrrrr}
\toprule
$I_0$ & log-RMSE & Pooled $\rho$ & Within-image $\rho$ & Case-mean $\rho$ & Admissible (\%)\\
\midrule
$1.50\times10^4$ & 0.953 & 0.832 & 0.834 & 0.879 & 98.81\\
$2.25\times10^4$ & 0.631 & 0.884 & 0.883 & 0.819 & 99.17\\
$3.50\times10^4$ & 0.439 & 0.932 & 0.925 & 0.886 & 99.55\\
$5.25\times10^4$ & 0.532 & 0.944 & 0.940 & 0.876 & 99.49\\
$8.00\times10^4$ & 0.732 & 0.951 & 0.947 & 0.850 & 99.49\\
\bottomrule
\end{tabular}
\end{table}

\begin{figure}[t]
\centering
\includegraphics[width=\linewidth]{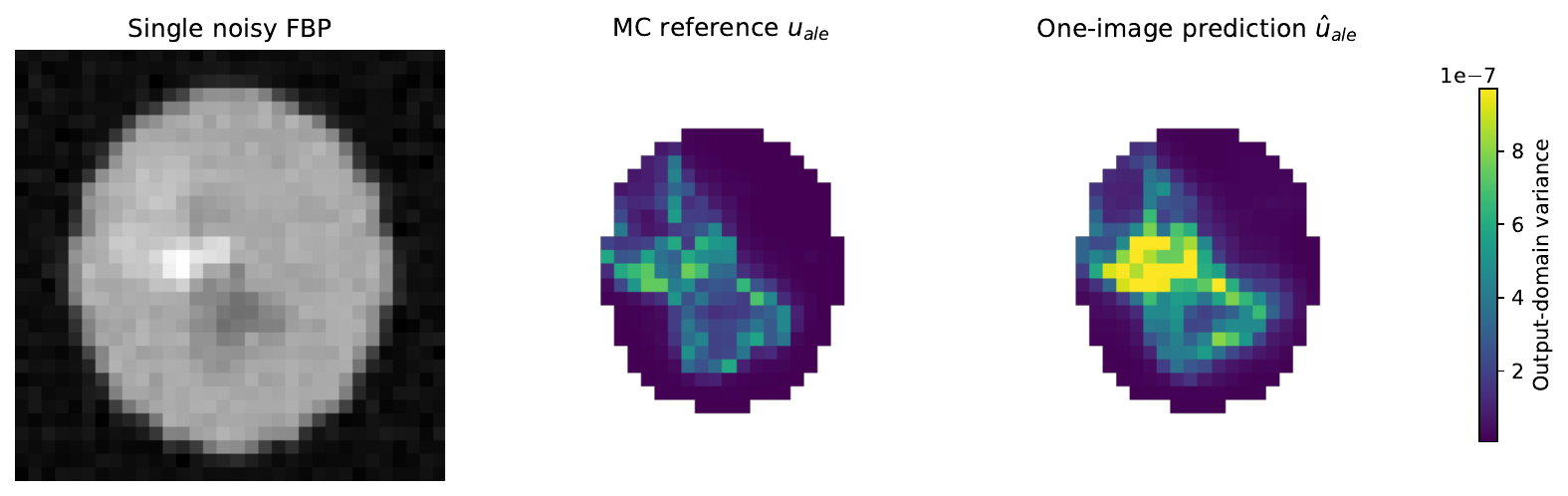}
\caption{Representative held-out object at $I_0=5.25\times10^4$. Left: the single noisy FBP supplied at inference. Middle: output-domain $u_{\mathrm{ale}}^{\mathrm{ref}}$ from 400 independent physical-noise realizations. Right: $\widehat u_{\mathrm{ale}}$ predicted from the single FBP. The variance maps share identical color limits.}
\label{fig:uale}
\end{figure}

Figure~\ref{fig:uale} shows a representative case selected by an objective rule: the case-dose pair whose within-image correlation is closest to the overall median. The prediction reproduces the main high-variance structures from a single FBP, although its peak amplitude is visibly higher than the MC reference, consistent with the residual magnitude discrepancy reflected in the nonzero log-RMSE.

These results establish only in-distribution prediction in this synthetic proof-of-concept. They do not establish out-of-distribution generalization, clinical performance, or ground-truth calibration of $u_{\mathrm{epi}}$, $\nu$, or $\beta$. The experiment was run once with a fixed seed and needs broader stability testing.

\section{Discussion and conclusion}
We underline the distinction between \emph{selecting} an NIG representative and \emph{identifying} one from additional information. The marginal Student-$t$ likelihood determines only the three coordinates $(\gamma,\alpha,c)$ and is invariant along the one-dimensional $(\beta,\nu)$ fiber. A KL term, reference prior, or evidence regularizer can favor a particular point on this fiber, but does not by itself provide an independent observation of the missing coordinate. PEEL instead introduces such information through the output-domain aleatoric variance measured from repeated physical-noise realizations. Once $u_{\mathrm{ale}}$ is supplied independently, $(\beta,\nu)$ follow uniquely from the algebraic relations in Eq.~(10). The sequential design is therefore essential: the Student-$t$ predictive model is first learned and frozen, and the aleatoric target is learned only afterward from frozen features, so that no aleatoric loss can alter the likelihood solution and no cross-loss weighting is required.

Also, we acknowledge the difference between recovering the spatial ordering of aleatoric uncertainty and reproducing its absolute magnitude. Across all five dose levels, the pooled and within-image Spearman correlations remained high, indicating that the one-image predictor reliably located relatively high- and low-variance regions. In contrast, the nonzero log-RMSE shows that some magnitude mismatch remained, as is also visible in Figure~3. Importantly, log-RMSE need not decrease monotonically with increasing dose, because it measures disagreement in the logarithm of the predicted and reference variances rather than reconstruction error itself. The lowest log-RMSE occurred at the middle dose, while the dose extremes showed larger residual calibration errors. At low dose, stronger measurement noise may make the output variance more difficult to predict, whereas at high dose the aleatoric variance becomes small, so even modest absolute discrepancies can correspond to appreciable relative errors, which are directly reflected in the log-domain metric. The case-mean correlations of 0.819--0.886 further indicate that the method preserved the ordering of overall uncertainty levels across held-out object--dose cases, although this proof-of-concept remains limited to in-distribution synthetic data, one fixed training seed, and Monte Carlo rather than clinical ground truth.

In conclusion, PEEL provides a physics-enabled route to identifiable evidential learning by combining likelihood-identifiable Student-$t$ coordinates with an independently measured aleatoric quantity. In this synthetic CT study, a network using only one noisy FBP at inference reproduced the spatial structure of a 400-repeat output-domain variance reference with strong correlation, while more than 98.8\% of evaluated pixels satisfied the admissibility condition required for algebraic recovery of the full NIG parameters. These results support the central premise that additional physical information can resolve NIG non-identifiability without a KL term, reference prior, evidence regularizer, or cross-loss coefficient. Future work should test robustness across random seeds, object distributions, acquisition geometries, reconstruction architectures, and real CT measurements, and should determine how well the identified uncertainty components generalize beyond the in-distribution setting.

\medskip
\paragraph{Author--AI Collaboration.}
Generative AI tools, including ChatGPT and Claude, were used to assist with formulation, drafting, simulation, and cross-checking. The author conceived the core idea and overall framework, directed and evaluated the AI-assisted work, and takes full responsibility for the content.

\end{document}